\documentclass[runningheads]{llncs}
\usepackage[breaklinks=true,letterpaper=true,colorlinks,citecolor=blue,bookmarks=false]{hyperref}
\usepackage{graphicx}
\usepackage{cite}
\usepackage{xcolor}
\usepackage{amsfonts}
\usepackage{amsmath}
\usepackage{caption}
\usepackage{subcaption}
\usepackage{blindtext}
\usepackage{hyperref}
\usepackage{cleveref}

\usepackage[resetlabels]{multibib}
\newcites{Supp}{References}
\begin{document}
\title{PAC Bayesian Performance Guarantees for Deep (Stochastic) Networks in Medical Imaging}
\titlerunning{PAC Bayesian Performance Guarantees in Medical Imaging}
%
\author{Anthony Sicilia\inst{1} \and Xingchen Zhao\inst{2} \and \\
Anastasia Sosnovskikh\inst{2} \and Seong Jae Hwang\inst{1,2}}


\institute{Intelligent Systems Program, University of Pittsburgh \and
Department of Computer Science, University of Pittsburgh \\
\email{\{anthonysicilia, xiz168, anastasia, sjh95\}@pitt.edu}}
\authorrunning{Sicilia et al.}
%
%
\maketitle              
\begin{abstract}
Application of deep neural networks to medical imaging tasks has in some sense become commonplace. Still, a “thorn in the side” of the deep learning movement is the argument that deep networks are prone to overfitting and are thus unable to generalize well when datasets are small (as is common in medical imaging tasks). One way to bolster confidence is to provide mathematical guarantees, or bounds, on network performance after training which explicitly quantify the possibility of overfitting. In this work, we explore recent advances using the PAC-Bayesian framework to provide bounds on generalization error for large (stochastic) networks.
While previous efforts focus on classification in larger natural image datasets (e.g., MNIST and CIFAR-10), we apply these techniques to both classification and segmentation in a smaller medical imagining dataset: the ISIC 2018 challenge set. 
We observe the resultant bounds are competitive compared to a simpler baseline, while also being more explainable and alleviating the need for holdout sets.
\end{abstract}
\section{Introduction}
Understanding the generalization of learning algorithms is a classical problem. Practically speaking, verifying whether a fixed method of inference will generalize may not seem to be a challenging task. Holdout sets are the tool of choice for most practitioners -- when sample sizes are large, we can be confident the measured performance is representative. In medical imaging, however, sample sizes are often small and stakes are often high. Thus, mathematical guarantees\footnote{Guarantees in this paper are probabilistic. Similar to confidence intervals, one should interpret with care: the guarantees hold with high probability prior to observing data.} on the performance of our algorithms are of paramount importance. Yet, it is not abundantly common to provide such guarantees in medical imaging research on deep neural networks; we are interested in supplementing this shortage.

A simple (but effective) guarantee on performance is achieved by applying a Hoeffding Bound to the error of an inference algorithm reported on a holdout set. In classification tasks, Langford provides a useful tutorial on these types of high probability bounds among others~\cite{langford2005tutorial}. The bounds are easily extended to any bounded performance metrics in general, and we use this methodology as a baseline in our own experimentation (Section~\ref{sec:experiments}). While effective, Hoeffding's Inequality falls short in two regards: (1) use of a holdout set requires that the model does not see all available data and (2) the practitioner gains no insight into \textit{why} the model generalized well. Clearly, both short-comings can be undesirable in a medical imaging context: (1) access to the entire dataset can be especially useful for datasets with rare presence of a disease and (2) understanding \textit{why} can both improve algorithm design and give confidence when deploying models in a clinical setting. Thus, we desire practically applicable bounds -- i.e., competitive with Hoeffding's Inequality -- which avoid the aforementioned caveats. 

Unfortunately, for deep neural networks, \textit{practically applicable} guarantees of this nature can be challenging to produce.
Traditional PAC bounds based on the Vapnik-Chervonenkis (VC) dimension~\cite{vapnik1971uniform, valiant1984theory, blumer1989learnability, shalev2014understanding} accomplish our goals to some extent, but require (much) more samples than parameters in our network to produce a good guarantee.\footnote{We (very roughly) estimate this by Thm.~6.11 of Shalev-Shwartz \& Ben-David~\cite{shalev2014understanding}. Bartlett et al.~\cite{bartlett2019nearly} provide tight bounds on VC dimension of ReLU networks. Based on these, the sample size must be magnitudes larger than the parameter count for a small generalization gap. See Appendix for additional details and a plot.} When our networks are large -- e.g., a ResNet-18 \cite{he2016deep} with more than 10M parameters -- our datasets are thus required to be unreasonably sized to ensure generalization, especially, in medical imaging contexts. In effect, these bounds are \textit{vacuous}; they are logically meaningless for the sample sizes we observe in practice. Specifically, vacuous describes any bound on error which is larger than 1, and therefore, gives us no new insight on a model's generalization ability because error rates lie in the range $[0,1]$. The term was coined by Dziugaite \& Roy~\cite{dziugaite2017computing} who observed that most PAC bounds on error for deep networks were in fact vacuous (when computed). The authors demonstrate a technique to compute non-vacuous bounds for some deep stochastic networks trained on standard datasets (e.g., MNIST) using the \textit{PAC-Bayesian} framework.\footnote{PAC-Bayes is attributed to McAllester~\cite{mcallester1999some}; also, Shawe-Taylor \& Williamson~\cite{shawe1997pac}.} As discussed later, the motivation for using PAC-Bayesian bounds is the hypothesis that stochastic gradient descent (SGD) trained networks generalize well when their solutions lie in large, flat minima.\footnote{Early formulations of this hypothesis are due to  Hochreiter \& Schmidhuber~\cite{hochreiter1997flat}.} Thus, these PAC-Bayes bounds give us exactly what we desire. They provide practically applicable guarantees on the performance of deep networks, while allowing the model access to all data and also giving insight on what network properties lead to good generalization in practice. To borrow an analogy from Arora~\cite{arora_blog}, this approach is ``prescriptive.'' It is similar to a doctor's orders to resolve high blood pressure: cut the salt, or in our case, look for flat minima.

In this context, the end goal of this paper is to validate whether the ``flat minima prescription''  -- with observed evidence in traditional computer vision -- may find similar success in explaining generalization for the small-data, non-traditional tasks that are common to medical imaging. Our contributions, in this sense, are primarily experimental. We demonstrate non-vacuous PAC-Bayesian performance guarantees for deep stochastic networks applied to the classification and segmentation tasks within the ISIC 2018 Challenge~\cite{codella2019skin}. Importantly, our results show that PAC-Bayesian bounds are competitive against Hoeffding's Inequality, offering a practical alternative which avoids the aforementioned caveats. We employ much the same strategies used by Dziugaite et al. \cite{dziugaite2017computing, dziugaite2020role} as well those used by P{\'e}rez-Ortiz et al. \cite{perez2020tighter}. With that said, our different setting yields novel experimental results and poses some novel challenges. Specifically, in segmentation, we compute non-vacuous bounds for a fully-sized U-Net using a medical imaging dataset with about 2.3K training samples. To our knowledge, for deep stochastic networks, we are the first to compute non-vacuous bounds in segmentation on such small datasets. Along the way, we offer some practical insights for the medical imaging practitioner including a (mathematically sound) trick to handle batch normalization layers in PAC-Bayes bounds and an experimental technique
to ``probe'' parameter space to learn about the generalization properties of a particular model and dataset. We hope these results promote continued research on the important topic of guarantees in medical imaging.
\section{PAC-Bayesian Theory and Generalization}
\label{sec:pacbayes}
\subsection{Formal Setup}
In the PAC-Bayes setting, we consider a hypothesis space $\mathcal{H}$ and a distribution $Q$ over this space. Specific to our context, $\mathcal{H} = \mathbb{R}^d$ will represent the space of deep networks with some fixed architecture, and $Q$ will be a distribution over $\mathcal{H}$. Typically, we will set $Q = \mathcal{N}(\mu, \Sigma)$, a multivariate normal. For some fixed space $\mathcal{Z} = \mathcal{X} \times \mathcal{Y}$, the hypothesis $h \in \mathcal{H}$ defines a mapping $x \mapsto h(x)$ with $x\in \mathcal{X}$, and $h(x) \in \mathcal{Y}$. Given a $[0,1]$-bounded loss function $\ell : \mathcal{H} \times \mathcal{Z} \to [0,1]$ and a distribution $D$ over $\mathcal{Z}$, the risk of $h$ is defined $\mathcal{R}_\ell(h, D) = \mathbf{E}_{(x,y) \sim D} \ \ell(h, (x,y))$.
Given instead a sample $S \sim D^m$ over $\mathcal{Z}$, the empirical risk is denoted $\hat{\mathcal{R}}_\ell(h, S)$ and is computed as usual by averaging.
In all of our discussions, the data distribution $D$ or sample $S$ is usually easily inferred from context. Therefore, we typically write $\mathcal{R}_\ell(h) = \mathcal{R}_\ell(h, D)$ and $\hat{\mathcal{R}}_\ell(h) = \hat{\mathcal{R}}_\ell(h, S)$. With these definitions, we are interested in quantifying the risk of a stochastic model.\footnote{Sometimes, in classification, this may be called the Gibbs classifier. Not to be confused with the ``deterministic,'' majority vote classifier. An insightful discussion on the relationship between risk in these distinct cases is provided by Germain et al.~\cite{germain2015risk}.} In the context of neural networks, one can imagine sampling the distribution $Q$ over $\mathbb{R}^d$ and setting the weights before performing inference on some data-point $x$. Often, we will refer to $Q$ itself as the stochastic predictor. The associated risk for such a model is defined as $\mathcal{R}_\ell(Q) = \mathbf{E}_{h \sim Q} \ \mathcal{R}_\ell(h)$
with $\hat{\mathcal{R}}_\ell(Q)$ similarly defined. Typically, we cannot exactly compute expectations over $Q$. For this reason, we also define $\hat{R}_\ell(\hat{Q})$ for a sample $\hat{Q} \sim Q^n$ as an empirical variant, computed by averaging.
The last components for any PAC-Bayesian bound come from two notions of the Kullback-Leibler (KL) divergence between two distributions $Q$ and $P$ written $\mathrm{KL}(Q || P)$ and defined as usual. For numbers $q, p \in [0,1]$, we write $\mathrm{kl}(q || p)$ as shorthand for the KL divergence between Bernoulli distributions parameterized by $q, p$. This is typically used to quantify the difference between the risk on a sample and the true risk. In the next section, we put the discussed pieces in play.
\subsection{The PAC-Bayesian Theorem}
The PAC-Bayesian theory begins primarily with the work of McAllester~\cite{mcallester1999some} with similar conceptualizations given by Shawe-Taylor \& Williamson~\cite{shawe1997pac}. Besides what is discussed in this section, for completeness, readers are also directed to the work of Catoni~\cite{catoni2007pac}, McAllester~\cite{mcallester2013pac}, Germain et al.~\cite{germain2009pac, germain2015risk}, and the primer by Guedj~\cite{guedj2019primer}. We start by stating the main PAC-Bayesian Theorem as given by Maurer~\cite{maurer2004note}. See also Langford \& Seeger~\cite{langford2001bounds} for the case where $\ell$ is a $01$ loss. 
\begin{theorem}
\label{thm:Maurer}
(Maurer) Let $\ell$ be a $[0,1]$-bounded loss function, $D$ be a distribution over $\mathcal{Z}$, and $P$ be a probability distribution over $\mathcal{H}$. Then, for $\delta \in (0,1)$
\begin{equation}\small
\label{eqn:Maurer_bound}
\underset{S \sim D^m}{\mathbf{Pr}} \left (\forall \ Q \  : \ \mathrm{kl}\left(\hat{\mathcal{R}}_\ell(Q) || \mathcal{R}_\ell(Q) \right) \leq \frac{\mathrm{KL}(Q || P) + \ln \frac{1}{\delta} + \ln \sqrt{4m} }{m} \ \right) \geq 1 - \delta.
\end{equation}
\end{theorem}
By way of Pinsker's inequality, the above may be loosened for the purpose of interpretation \cite{germain2015risk}
\begin{equation}\small
\label{eqn:relaxation}
    \mathcal{R}_\ell(Q) \leq \hat{\mathcal{R}}_\ell(Q) + \sqrt{\frac{\mathrm{KL}(Q || P) + \ln \frac{1}{\delta} + \ln \sqrt{4m} }{2 m}}.
\end{equation}
In Section~\ref{sec:experiments}, we compute a much tighter formulation of the bound given in Eqn.~\eqref{eqn:relaxation} which handles the term $\mathrm{kl}\left(\hat{\mathcal{R}}_\ell(Q) || \mathcal{R}_\ell(Q) \right)$ directly. We provide the derivation of this bound in the Appendix. Various insights and results used to build the final bound are of course due to Langford \& Caruana~\cite{langford2002not}; Dziugaite et al.~\cite{dziugaite2017computing, dziugaite2020role}; and P{\'e}rez-Ortiz et al.~\cite{perez2020tighter} who have all computed similar (or identical) bounds on stochastic neural networks before us. In Section~\ref{sec:experiments}, we compute this bound for classification and segmentation tasks. For classification, we take $\ell$ to be the 01-loss defined $\ell_{01}(h, (x,y)) = \mathbf{1}[h(x) \neq y]$ where $\mathbf{1}$ is the indicator function. Precisely, $\mathcal{R}_{\ell_{01}}(h)$ is equal to 1 minus the accuracy. For segmentation, we pick $\ell$ to be $\ell_\mathrm{DSC}(h, (x,y)) = 1 - \mathrm{DSC}(h, (x,y))$ where $\mathrm{DSC}$ is the $[0,1]$-valued Dice similarity coefficient. These upperbounds trivially yield corresponding lowerbounds for both the accuracy and the Dice similarity coefficient, respectively.
\newline\textbf{Selecting the Prior.} Often $P$ is referred to as the prior and $Q$ as the posterior. Still, it is important to note that no restriction on the form of $P$ and $Q$ is required (e.g., as it is in Bayesian parameter estimation). 
What is required is that $P$ be fixed before observing the sample $S$ that is used to compute the bound. Albeit, $P$ can depend on $D$ and samples independent of $S$. In fact, it is not uncommon for the prior $P$ to be \textbf{data-dependent}.\footnote{For example, see Ambroladze et al.~\cite{ambroladze2007tighter}, Parrado-Hern{\'a}ndez et al.~\cite{parrado2012pac}, P{\'e}rez-Ortiz et al.~\cite{perez2020tighter}, and Dziugaite et al.~\cite{dziugaite2020role,dziugaite2018data}.} That is, $P$ may be trained on a sample which is disjoint from that which is used to compute the bound; i.e., disjoint from $S$. On the other hand, the bound holds for all posteriors $Q$ regardless of how $Q$ is selected. So, the datasets used to train $Q$ and $P$ may actually intersect. All in all, we can train $Q$ with all available data without violating assumptions. We must only ensure the datasets used to train $P$ and compute the bound do not intersect. In effect, we avoid the first caveat of Hoeffding's Inequality.
\newline\textbf{Interpretation.} The bound also offers insight into \textit{why} the model generalizes. Intuitively, we quantify the complexity of the stochastic predictor $Q$ in so much as it deviates from some prior knowledge we have on the solution space (e.g., from the data-dependent prior). This is captured in the term $\mathrm{KL}(Q||P)$. Dziugaite \& Roy~\cite{dziugaite2017computing} also relate PAC-Bayesian bounds to the flat-minima hypothesis. 
To understand their observation, we consider the case where $Q$ is a normal distribution $\mathcal{N}(\mu, \sigma^2 I)$ with $\sigma$ a constant and $I$ the identity matrix. The model $Q$ on which we bound the error is stochastic: each time we do inference, we sample from $\mathcal{N}(\mu, \sigma^2 I)$. Because the distribution has some variance (dictated by $\sigma$), we sample network weights in a region \textit{around} the mean. Thus, when performance of the stochastic model $Q$ is good, there must be a non-negligible area around the mean where most networks perform well, i.e., a flat minimum around the mean. We know the variance is non-negligible in the posterior network $Q$ because a small upperbound implies small KL divergence with the prior $P$ which itself has non-negligible variance (we pick this value). So, to reiterate, small KL divergence and small empirical risk imply a flat-minimum of appropriate size around the mean of $Q$. In this sense the bound is explainable: $Q$ generalizes well because it does not deviate much from prior knowledge and it lies in a flat minimum.
\newline\textbf{Additional Context.}
Dziugaite \& Roy~\cite{dziugaite2017computing} provide a nice synopsis of the history behind the flat-minima hypothesis including the work of Hochreiter \& Schmidhuber~\cite{hochreiter1997flat}, Hinton \& Van Camp~\cite{hinton1993keeping}, Baldassi et al.~\cite{baldassi2015subdominant, baldassi2016unreasonable}, Chaudhari et al.~\cite{chaudhari2016entropy}, and Keskar et al.~\cite{keskar2016large}. Since then, large scale empirical studies  -- e.g., Jiang et al.~\cite{jiang2019fantastic}; Dziugaite, Drouin, et al. \cite{dziugaite2020search} -- have continued to indicate that measures of sharpness of the minimum may be good indicators of neural network generalization in practice. For completeness, we also point out some other theoretically plausible indicators of deep network generalization. These include small weight norms -- e.g., Bartlett~\cite{bartlett1997valid, bartlett1998sample};  Neyshabur et al.~\cite{neyshabur2014search, neyshabur2017exploring} -- and the notion of algorithmic stability proposed by Bousquet \& Elisseeff~\cite{bousquet2002stability} which focuses instead on the SGD algorithm -- e.g., Hardt et al.~\cite{hardt2016train}; Mou et al. \cite{mou2018generalization}; Kuzborskij \& Lampert~\cite{kuzborskij2018data}.
\section{Experiments}
\label{sec:experiments}
In this section, we first evaluate PAC-Bayesian bounds within a self-bounded learning setting. Specifically, a self-bounded learner must both learn and provide a guarantee using the same dataset.\footnote{See Freund~\cite{freund1998self} or Langford \& Blum~\cite{langford2003microchoice}.} As noted, providing guarantees with our trained networks can bolster confidence in small data regimes. We compare the PAC-Bayesian bounds discussed in Section~\ref{sec:pacbayes} to a simple baseline for producing performance guarantees: application of Hoeffding's Inequality to a holdout set.\footnote{We provide additional details on this procedure in the Appendix.} We show PAC-Bayesian bounds are competitive with Hoeffding's Inequality, while also alleviating some caveats discussed in the previous sections. This result (in medical imaging) compliments those previously shown on natural image datasets by P{\'e}rez-Ortiz et al. \cite{perez2020tighter}. Specifically, we compute bounds on the \textbf{Lesion Segmentation} and the \textbf{Lesion Classification Tasks} in the ISIC 2018 Skin Lesion Challenge Dataset \cite{codella2019skin}. The data in these sets used for training (and bound computation) totals 2.3K and 9K labeled examples, respectively, which is much smaller than previous computation \cite{dziugaite2017computing, perez2020tighter, dziugaite2020role} using MNIST~\cite{lecun-mnisthandwrittendigit-2010} or CIFAR-10~\cite{krizhevsky2009learning}.\footnote{These datasets have 60K and 50K labeled examples, respectively} Our second contribution in this section comes from tricks and tools which we hope prove useful for the medical imaging practitioner. We demonstrate an experiment to probe the loss landscape using PAC-Bayesian bounds and also devise a strategy to handle batch normalization layers when computing PAC-Bayesian bounds. Our code is available at: \url{https://github.com/anthonysicilia/PAC-Bayes-In-Medical-Imaging}.
\subsection{Setup}
\begin{figure}[t!]
    \centering
    \begin{subfigure}[b]{0.24\textwidth}
        \includegraphics[width=\textwidth, trim={15 0 10 0}]{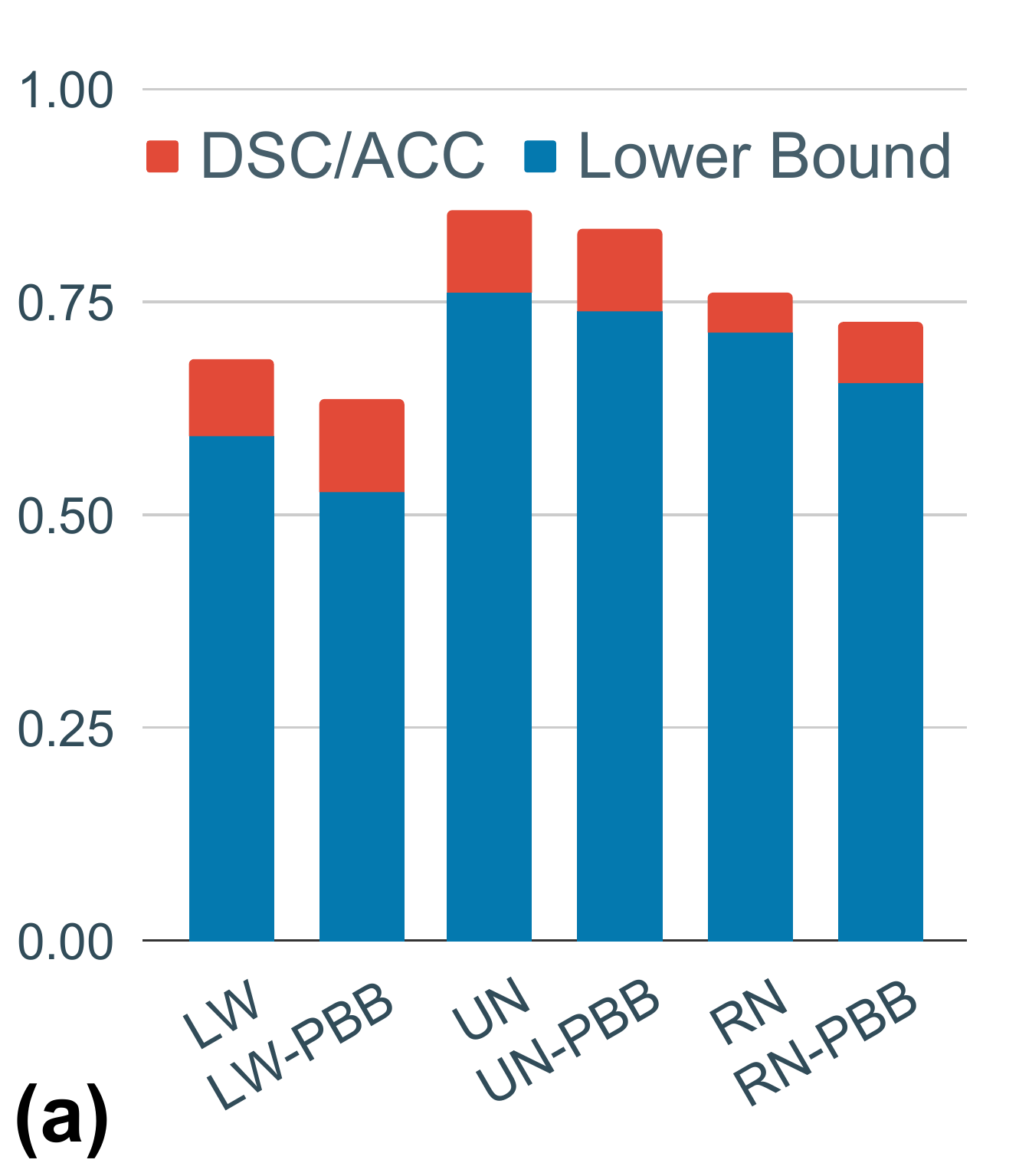}
    \end{subfigure}
    \hfill
    \begin{subfigure}[b]{0.24\textwidth}
        \includegraphics[width=\textwidth, trim={15 0 10 10}]{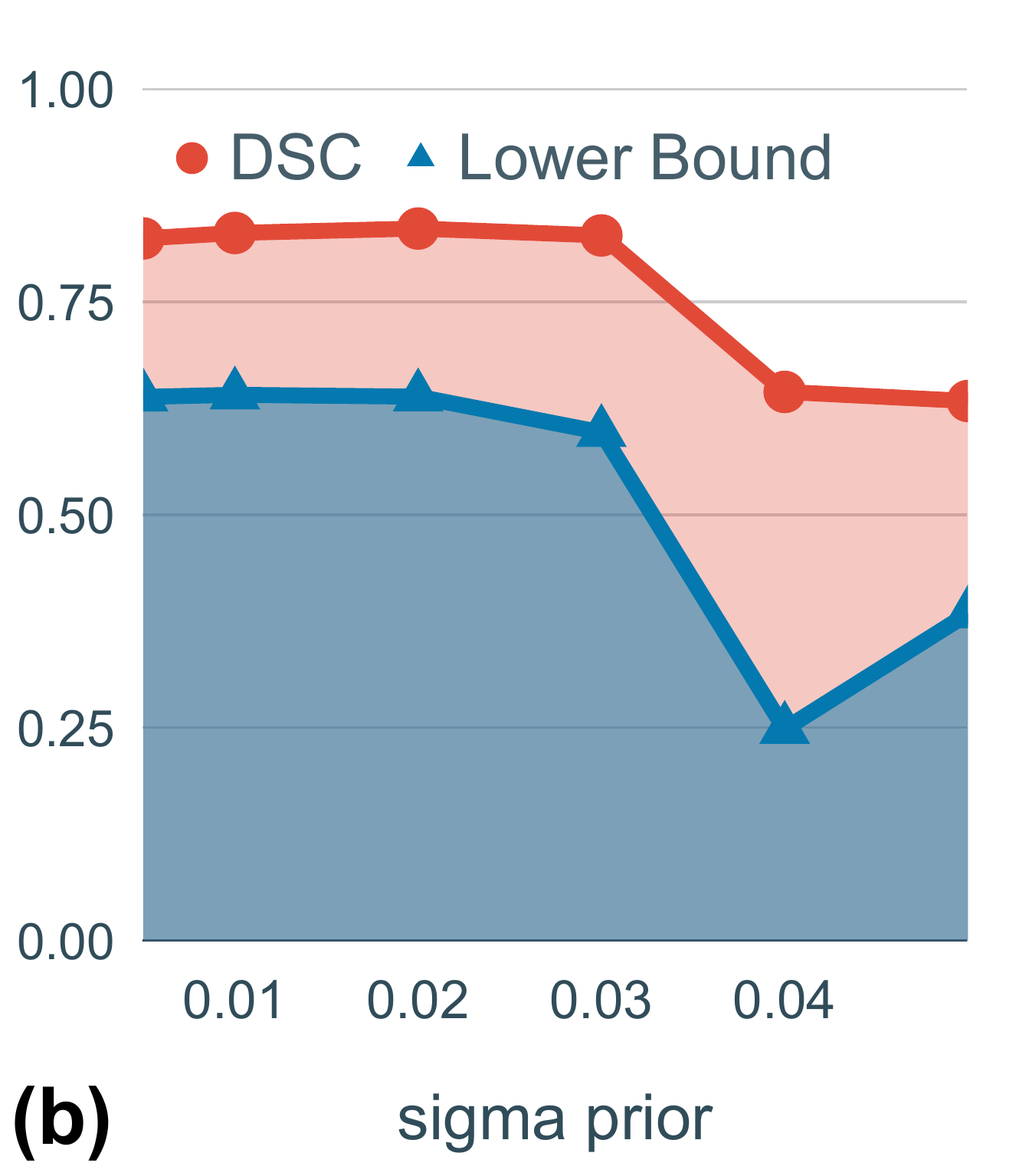}
    \end{subfigure}
    \hfill
    \begin{subfigure}[b]{0.24\textwidth}
        \includegraphics[width=\textwidth, trim={15 0 10 10}]{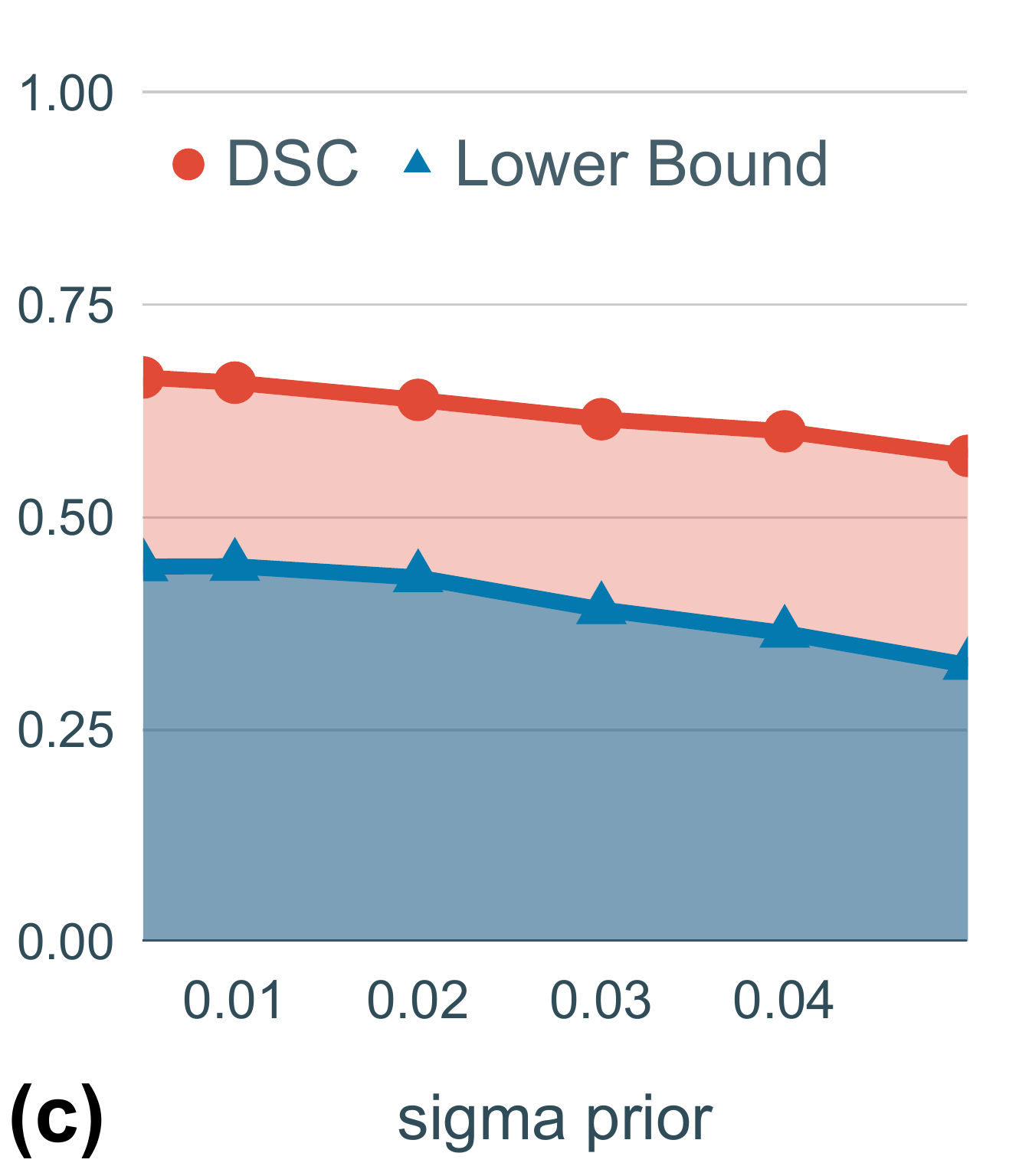}
    \end{subfigure}
    \hfill
    \begin{subfigure}[b]{0.24\textwidth}
        \includegraphics[width=\textwidth, trim={15 0 5 10}]{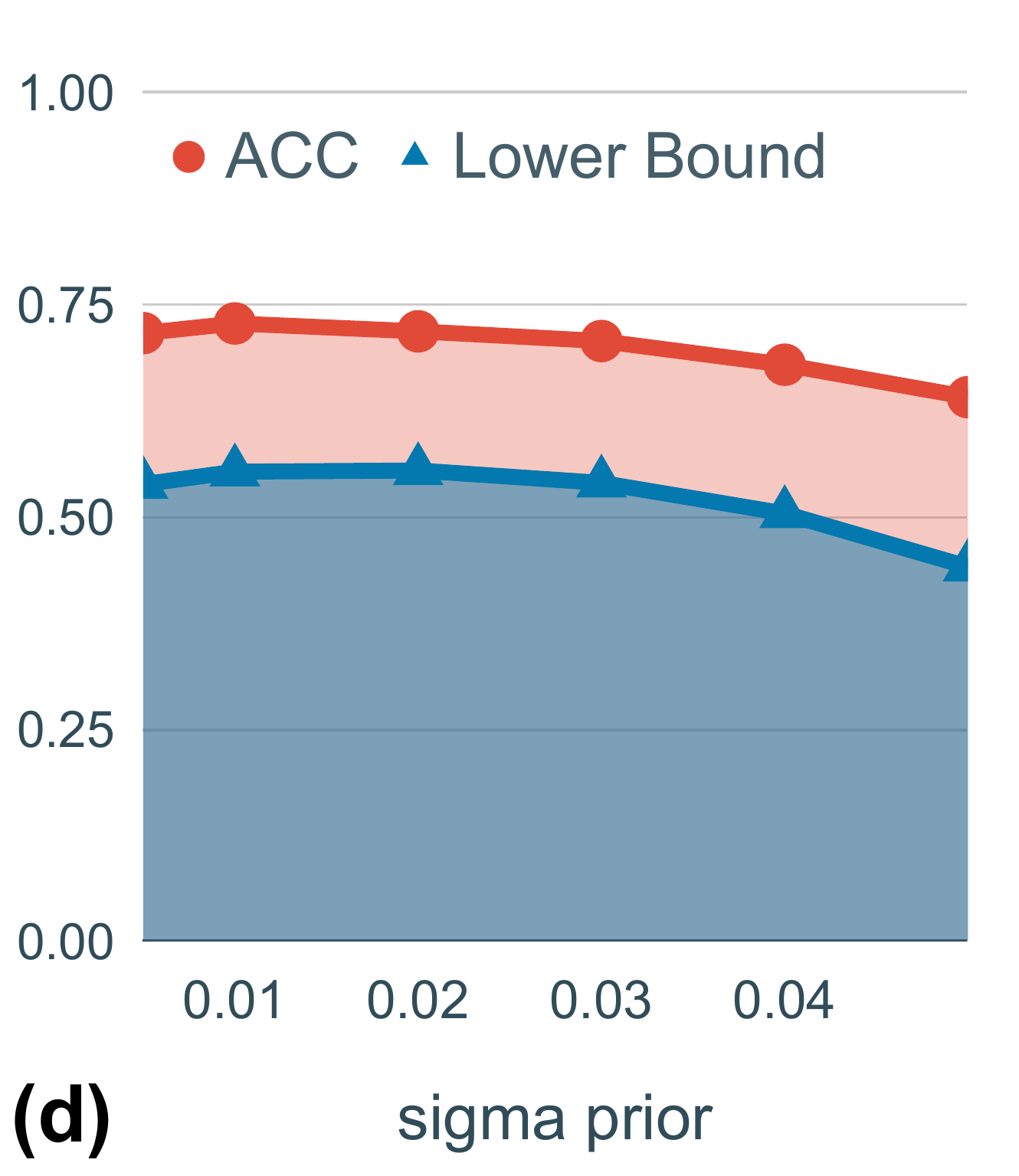}
    \end{subfigure}
    \caption{\textbf{(a)} DSC/ACC (red) and lowerbounds (blue). \textbf{(b,c,d)} Modulation of prior variance for U-Net \textbf{(b)} and LW \textbf{(c)} and ResNet-18 \textbf{(d)}.}
\label{fig:main}
\end{figure}

\textbf{Models.} For segmentation, we use U-Net (UN) \cite{ronneberger2015u} and a light-weight version of U-Net (LW) with 3\% of the parameters and no skip-connections. For classification, we use ResNet-18 (RN) \cite{he2016deep}. Probabilistic models use the same architecture but define a multivariate normal distribution over network weights $\mathcal{N}(\mu, \Sigma)$ with a diagonal covariance matrix $\Sigma$. The distribution is sampled to do inference.
\newline\textbf{Losses.} For segmentation, we train using the common Dice Loss \cite{milletari2016v} which is a smooth surrogate for $1 - DSC$. For classification, we use the negative log-likelihood. Probabilistic models used modified losses we describe next.
\newline\textbf{Training Probabilistic Models.} Recall, in the PAC-Bayes setting we define both the prior $P$ and the posterior $Q$ as both are needed to compute bounds and $Q$ is needed for inference. The prior $P$ is a probabilistic network defined by $\mathcal{N}(\mu_\mathrm{p}, \sigma_\mathrm{p}^2I)$ where $I$ is the identity matrix and $\sigma_\mathrm{p}$ is a constant. In this text, we use a data-dependent prior unless otherwise noted (see Section~\ref{sec:pacbayes}). To pick the prior, $\mu_\mathrm{p}$ is learned by performing traditional optimization on a dataset disjoint from that which is used to compute the bound (see \textbf{Data Splits}). The parameter $\sigma_\mathrm{p} = 0.01$ unless otherwise noted. The posterior $Q$ is initialized identically to $P$ before it is trained using PAC-Bayes with Backprop (PBB) as proposed by Rivasplata et al.~\cite{rivasplata2019pac, perez2020tighter}. This training technique may be viewed as (mechanically) similar to Bayes-by-Backprop (BBB) \cite{blundell2015weight}. In particular, it uses a re-parameterization trick to optimize a probabilistic network through SGD. Where PBB and BBB differ is the motivation, and subsequently, the use of PAC-Bayes upperbounds as the objective to optimize.\footnote{See P{\'e}rez-Ortiz et al. \cite{perez2020tighter} for more detailed discussion.} Note, PAC-Bayes bounds are valid for all $[0,1]$-bounded losses, and thus, are valid for the Dice Loss or normalized negative log-likelihood. The upperbound used for our PBB objective is the Variational bound of Dziugaite et al.~\cite{dziugaite2020role}.
\newline\textbf{Probabilistic Batch Norm Layers.} While generally each weight in the probabilistic networks we consider is sampled independently according to a normal distribution, batch norm layers\footnote{We refer here to both the running statistics and any learned weights.} must be handled with special care. We treat the parameters of batch norm layers as point mass distributions. Specifically, the parameter value has probability 1 and 0 is assigned to all other values. The posterior distribution for these parameters is made identical to the prior by ``freezing'' the posterior batch norm layers during training and inference. In effect, we avoid sampling the means and variances in our batch norm layers, and importantly, the batch norm layers do not contribute to the KL-divergence computation. In the Appendix, we provide a derivation to show this strategy is (mathematically) correct; it relies primarily on the independence of the weight distributions.
\newline\textbf{Optimization Parameters.} Optimization is done using SGD with momentum set to 0.95. For classification, the batch size is 64, and the initial learning rate is 0.5. For segmentation, the batch size is 8, and the initial learning rate is 0.1 for LW and 0.01 for U-Net. All models are initialized at the same random location and are trained for 120 epochs with the learning rate decayed by a factor of 10 every 30 epochs. In the PAC-Bayesian setting, the data-dependent prior mean $\mu_\mathrm{p}$ is randomly initialized (as other models) and trained for 30 epochs. The posterior $Q$ is initialized at the prior $P$ and trained for the remaining 90 epochs. The learning rate decay schedule is not reset for posterior training. 
\newline\textbf{Bound Details.} Note, in all cases, PAC-Bayes bounds are computed using a data-dependent prior. Bounds are computed with 95\% confidence ($\delta=0.05$) with data sample size given in the next section. For PAC-Bayes bounds, the number of models sampled is either 1000 (in Fig.~\ref{fig:main}a) or 100 (in Fig.~\ref{fig:main}b,c,d).
\newline\textbf{Data Splits for Self-Bounded Learning.} Each method is given access to a base training set (90\% of the data) and is expected to both learn a model and provide a performance guarantee for this model when applied to unseen data. To evaluate both the model and performance guarantee in a more traditional fashion, each method is also tested on a final holdout set (10\% of the data) which no model sees. Splits are random but identically fixed for all models.
For probabilistic networks trained using PBB, we split the base training data into a 50\%-prefix set\footnote{See Dziugaite et al.~\cite{dziugaite2020role} who coin the term ``prefix.''} used to train the prior and a disjoint 50\%-bound set. Both the prefix-set and bound-set are used to train the posterior, but recall, the PAC-Bayes bound can only be computed on the bound-set (Section~\ref{sec:pacbayes}). For the baseline non-probabilistic networks, we instead train the model using a traditional training set (composed of 90\% of the base training set) and then compute a guarantee on the model performance (i.e., a Hoeffding bound) using an independent holdout set (the remaining 10\% of the base training set). In this sense, all models are on an equal footing with respect to the task of a self-bounded learner. All models have access \textit{only} to the base training set for both training and computation of valid performance guarantees. In relation to Fig.~\ref{fig:main}, performance metrics such as DSC are computed using the final holdout set. Lowerbound computation and training is done using the base training set.
\subsection{Results}
\textbf{Comparison to Hoeffding's Inequality.}
As promised, we observe in Fig.~\ref{fig:main}a performance guarantees by both bounds are comparable and
performance on the final holdout set is also comparable. Hoeffding's Inequality does have a slight advantage with regards to these metrics, but as mentioned, PAC-Bayes bounds possess some desirable qualitative properties which make it an appealing alternative. For PAC-Bayes bounds the posterior $Q$ sees all training data, while for the Hoeffding Bound, one must maintain an unseen holdout set to compute a guarantee. Further, we may explain the generalization of the PBB trained model through our interpretation of the PAC-Bayes bound (see Section~\ref{sec:pacbayes}). These qualities make PAC-Bayes appealing in medical imaging contexts where explainability is a priority and we often need to maximize the utility of the training data.
\newline\textbf{Flat Minima and Their Size.} As discussed, the application of PAC-Bayesian bounds may be motivated by the flat minima hypothesis (see Section~\ref{sec:pacbayes}). We explore this idea in Fig.~\ref{fig:main}b,c,d by modulating the prior variance $\sigma_\mathrm{p}$ across runs. Informally, these plots can give us insight into our loss landscape. The reasonably tight lowerbounds -- which are slightly looser than in Fig.~\ref{fig:main}a only due to fewer model samples -- imply small KL-Divergence and indicate the prior and posterior variances are of a similar magnitude. Likewise, the difference between the prior and posterior means should not be too large, relative to the variance. A fixed random-seed ensures priors are identical, so each data-point within a plot should correspond to roughly the same location in parameter space; i.e., we will assume we are analyzing the location of a single minimum.\footnote{Notice, another approach might be to the fix the posterior mean at the result of, say, the run with $\sigma_\mathrm{p}=0.01$ and then modulate the variance from this fixed location. We are not guaranteed this run will be near the center of a minimum, and so, may underestimate the minimum's size by this procedure. Our approach, instead, allows the center of the posterior to change (slightly) when the variance grows.} For U-Net, we see stable performance and a sudden drop as the prior variance grows. Before the drop at $\sigma_\mathrm{p}=0.04$, consistently high DSC and a high lowerbound indicate the network solution lies in a flat minimum (as discussed in Section~\ref{sec:pacbayes}). So, we may conclude a flat minimum proportional in size to $\sigma_\mathrm{p}=0.03$. For LW and ResNet-18, we instead see consistent performance degradation as the prior variance grows. For these networks, the minima may not be as flat. Informally, such sensitivity analysis can tell us ``how flat'' the minima are for a particular network and dataset as well as ``how large.'' Practically, information like this can be useful to practitioners interested in understanding the generalization ability of their models. Namely, it is hypothesized ``larger'' flat minima lead to better generalization because less precision (fewer bits) is required to specify the weights \cite{hochreiter1997flat}.
\section{Conclusion}
As a whole, our results show how PAC-Bayes bounds can be practically applied in medical imaging contexts -- where theoretical guarantees (for deep networks) would appear useful but not commonly discussed. With this said, we hope for this paper to act primarily as a conversation starter. At 2.3K examples, the segmentation dataset we consider is still larger than commonly available in some application domains (e.g., neuroimaging applications). It remains to be considered how effective these bounds can be in ultra-low resource settings. 
\newline \newline
\textbf{Acknowledgment.} This work is supported by the University of Pittsburgh Alzheimer Disease Research Center Grant (P30 AG066468).

%
%
\bibliographystyle{splncs04}
\bibliography{references}

\title{PAC Bayesian Performance Guarantees for Deep (Stochastic) Networks in Medical Imaging: Supplementary Material}
\titlerunning{PAC Bayesian Performance Guarantees in Medical Imaging}
%
\author{Anthony Sicilia\inst{1} \and Xingchen Zhao\inst{2} \and \\
Anastasia Sosnovskikh\inst{2} \and Seong Jae Hwang\inst{1,2}}

\institute{Intelligent Systems Program, University of Pittsburgh \and
Department of Computer Science, University of Pittsburgh \\
\email{\{anthonysicilia, xiz168, anastasia, sjh95\}@pitt.edu}}
\authorrunning{Sicilia et al.}
%
%
\maketitle              

\begin{appendix}
\section{Bound Computation}

\setcounter{theorem}{1}
We present here derivation of the bound we compute in Section~3. One option would be to apply Pinsker's inequality or slightly more complicated manipulations which produce tighter results when the KL divergence is small (e.g., see the Variational bound proposed by Dziugaite et al.~\citeSupp{dziugaite2020role2} and the Quadratic bound proposed by Rivasplata et al.~\citeSupp{rivasplata2019pac2, perez2020tighter2}). Unfortunately, these solutions tend to yield looser bounds as compared to other methods. As is common, we instead take the route which uses an inversion of the binary KL divergence:
\begin{equation}
\label{thm:inv_kl}
    \mathrm{kl}^{-1}(q || \varepsilon ) = \max \{p \in [0,1] \ \mid \ \mathrm{kl}(q || p) \leq \varepsilon\}.
\end{equation}
There are a few important properties to note about $\mathrm{kl}^{-1}$. First, by its design, it has the property that $p$ which satisfies $\mathrm{kl}(q || p) \leq \epsilon$ is bounded above by $\mathrm{kl}^{-1}(q || \epsilon)$. Taking $p = \mathcal{R}_\ell(Q)$ and $\epsilon$ to be the upperbound in (main text) Thm.~1, we may appropriately bound our quantity of interest. Second, although in practice $\mathrm{kl}^{-1}$ cannot be computed exactly, it can be approximated to arbitrary precision using the bisection method because the function $f(p) = \varepsilon - \mathrm{kl}(q || p)$ has a single root\footnote{For more on this fact, we like the discussion on $\mathrm{kl}^{-1}$ in the work of Reeb et al.~\citeSupp{reeb2018learning}} in the interval $[q, 1]$.

Next, we must also empirically estimate $\hat{\mathcal{R}}_\ell(Q)$, and, account for this estimate in our bound. This much can be accomplished by the sample convergence bound given by Langford and Caruana~\citeSupp{langford2002not2}. See also other works of Langford for more detailed exposition~\citeSupp{langford2002quantitatively, langford2005tutorial2}. 

\begin{theorem}
\label{thm:lang}
For all distributions $Q$, for any $\delta \in (0,1)$
\begin{equation}
    \underset{\hat{Q} \sim Q^n}{\mathbf{Pr}} \left ( \mathrm{kl}\left(\hat{\mathcal{R}}_\ell(\hat{Q}) || \hat{\mathcal{R}}_\ell(Q) \right) \leq \frac{\ln \frac{2}{\delta}}{n} \ \right) \geq 1 - \delta.
\end{equation}
\end{theorem}
Using iterated application of the function $\mathrm{kl}^{-1}$, it is straightforward to combine Thm.~1 (in main text) with Thm.~\ref{thm:lang} to arrive at a computable bound on $\mathcal{R}_\ell(Q)$ as given below
\begin{equation}
\label{eqn:final_bound}
\begin{split}
    \mathcal{R}_\ell(Q) & \leq \mathrm{kl}^{-1} \left (q \ || \ m^{-1}\left (\mathrm{KL}(Q || P) + \ln \tfrac{1}{\delta} + \ln \sqrt{4m} \ \right ) \right) \\ 
    & \text{where} \quad q = \mathrm{kl}^{-1} \left ( \hat{\mathcal{R}}_\ell(\hat{Q}) \ || \ n^{-1}\ln \tfrac{2}{\delta} \right).
\end{split}
\end{equation}
This uses the fact that $\mathrm{kl}^{-1}$ is monotone increasing in its first argument which is easily observed by computing its partial derivatives as in \citeSupp{reeb2018learning}.
\section{VC Dimension Bounds}

\begin{figure}[h!]
\centering
    \includegraphics[width=0.8\textwidth]{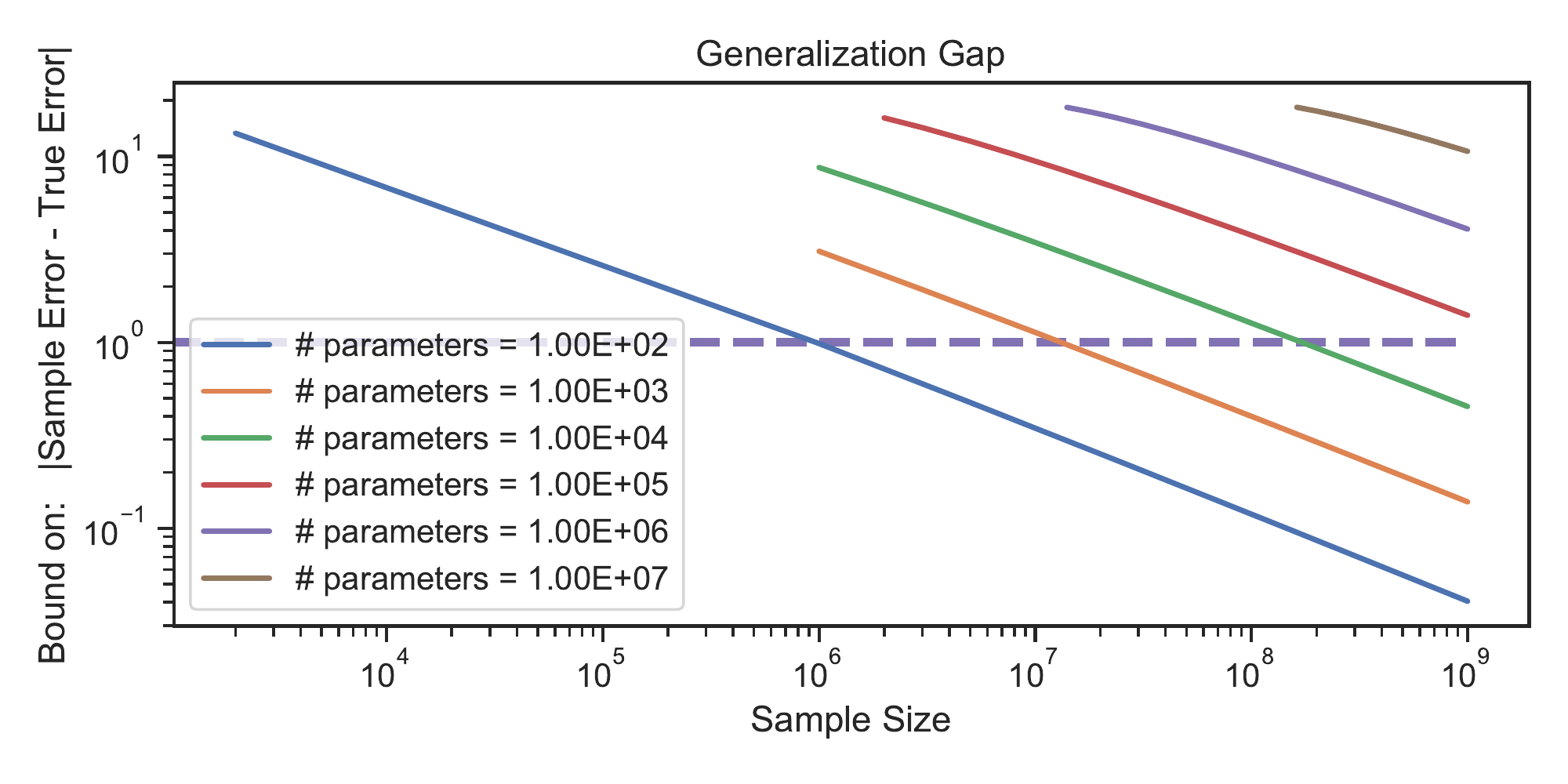}
    \caption{Bounds on generalization gap using VC Dimension}
\label{fig:main2}
\end{figure}

Fig.~\ref{fig:main2} shows the bounds on generalization gap using VC Dimension using Thm.~6.11 of \citeSupp{shalev2014understanding2} with $\delta=0.05$. We estimate network VC Dimension as $W \log W$ where $W$ is the parameter count \citeSupp{bartlett2019nearly2}; this ignores the \# of layers and constant factors. The bound is logically meaningful (i.e., non-vacuous) below the dotted line.
\section{Derivation for Probabilistic Batch Norm Layers}
For prior $P$ and posterior $Q$, let $P_i$ be the distribution of the $i$th parameter in the prior network architecture and $Q_i$ be similarly defined for the posterior network. For the posterior parameters which are normally distributed, let $\mu$ denote the learned mean and $\Sigma$ denote the learned (diagonal) co-variance matrix. As in Section~3, let $\mu_\mathrm{p}$ be the learned mean for the prior and $\sigma_\mathrm{p}^2$ be the fixed prior variance. Finally, let $\delta$ be the Dirac measure and let $\Theta_\mathrm{p}$ / $\Theta$ denote the batch normalization parameters learned for the prior $P$ / posterior $Q$. Then, the KL-divergence expands as follows
\begin{equation}
    \begin{split}
        & \mathrm{KL}(Q || P) = \sum\nolimits_i \mathrm{KL}(Q_i|| P_i) \quad (\text{independence of parameter sampling}) \\
        & = \mathrm{KL}(\mathcal{N}(\mu, \Sigma)|| \mathcal{N}(\mu_\mathrm{p}, \sigma^2_\mathrm{p} I)) + \mathrm{KL}(\delta_{\Theta} || \delta_{\Theta_\mathrm{p}}) \quad (\text{collecting terms}).
    \end{split}
\end{equation}
In our experimental setup, the term $\mathrm{KL}(\delta_{\Theta} || \delta_{\Theta_\mathrm{p}}) = \mathrm{KL}(\delta_{\Theta_\mathrm{p}} || \delta_{\Theta_\mathrm{p}}) = 0$ because we fix the batch normalization parameters during training of the posterior and initialize the posterior at the prior. Thus, this approach yields a simple way to handle batch normalization parameters without sampling. On the other hand, sampling of normalization parameters could lead to instability and would require custom implementations in most deep learning libraries. 

\section{Application of Hoeffding's Inequality}
For any sequence of independent random variables $(X_i)_{i=1}^n$ with each $X_i$ bounded in the range $[0,1]$, Hoeffding's Inequality states
\begin{equation}
    \mathbf{Pr}\left( \ \left | n^{-1} \sum\nolimits_i X_i - \mathbf{E} \left [n^{-1} \sum\nolimits_i X_i \right ] \right | \geq \epsilon \right) \leq 2\exp \left ( -2n\epsilon^2\right).
\end{equation}
For any $\delta \in (0,1)$, setting $\delta = 2\exp \left ( -2n\epsilon^2\right)$ implies $\epsilon = \sqrt{\log (2 / \delta) / (2n)}$. Thus, 
\begin{equation}
    \mathbf{Pr}\left( \ \left | n^{-1} \sum\nolimits_i X_i - \mathbf{E} \left [n^{-1} \sum\nolimits_i X_i \right ] \right | \geq \sqrt{\log (2 / \delta) / (2n)} \ \right) \leq \delta.
\end{equation}
Taking the compliment event and manipulating terms yields
\begin{equation}
\label{eqn:final_hoeff}
    \mathbf{Pr}\left( \mathbf{E} \left [n^{-1} \sum\nolimits_i X_i \right ]  \leq n^{-1} \sum\nolimits_i X_i + \sqrt{\log (2 / \delta) / (2n)} \ \right) \geq 1 - \delta.
\end{equation}
When a model $h$ is fixed and then a dataset $S = (x_i, y_i)_{i=1}^n$ is independently drawn from $D^n$, the sequence $(\mathbf{1}[h(x_i) \neq y_i])_{i=1}^n$ meets the assumptions required by Hoeffding's Inequality. Further, we have that the average of this sequence is the error rate (i.e., 1 minus the accuracy). Thus, Eq.~\eqref{eqn:final_hoeff} may be used to lowerbound the accuracy as is done in Section~3.
\end{appendix}

\bibliographystyleSupp{splncs04}
\bibliographySupp{references}
\end{document}